\documentclass[11pt]{article}

\usepackage[final]{acl}

\usepackage{times}
\usepackage{latexsym}
\usepackage{amsmath, amssymb}
\usepackage[linesnumbered,ruled,vlined]{algorithm2e}

\usepackage{booktabs}
\usepackage{multirow}
\usepackage[T1]{fontenc}
\usepackage{subcaption}

\usepackage[utf8]{inputenc}

\usepackage{microtype}

\usepackage{inconsolata}

\usepackage{graphicx}

\title{SINdex: \textbf{S}emantic \textbf{IN}consistency Index for Hallucination Detection in LLMs}

\author{
 \textbf{Samir Abdaljalil\textsuperscript{1}\thanks{Corresponding Authors: sabdaljalil@tamu.edu, hkurban@hbku.edu.qa}},
 \textbf{Hasan Kurban\textsuperscript{2}\footnotemark[1]},
 \textbf{Parichit Sharma\textsuperscript{3}},\\
 \textbf{Erchin Serpedin\textsuperscript{1}},
 \textbf{Rachad Atat\textsuperscript{4}}
\\
\\
 \textsuperscript{1}Electrical and Computer Engineering, Texas A\&M University, College Station, TX USA,\\
  \textsuperscript{2}College of Science and Engineering, Hamad Bin Khalifa University, Doha, Qatar,\\
 \textsuperscript{3}Department of Computer Science, Luddy School of Informatics, Bloomington, Indiana, USA, \\
 \textsuperscript{4}Department of Computer Science and Mathematics, Lebanese American University,
Beirut, Lebanon 
}

\begin{document}
\maketitle
\begin{abstract}
Large language models (LLMs) are increasingly deployed across diverse domains, yet they are prone to generating factually incorrect outputs—commonly known as “hallucinations.” Among existing mitigation strategies, uncertainty-based methods are particularly attractive due to their ease of implementation, independence from external data, and compatibility with standard LLMs. In this work, we introduce a novel and scalable uncertainty-based semantic clustering framework for automated hallucination detection. Our approach leverages sentence embeddings and hierarchical clustering alongside a newly proposed inconsistency measure, SINdex, to yield more homogeneous clusters and more accurate detection of hallucination phenomena across various LLMs. Evaluations on prominent open- and closed-book QA datasets demonstrate that our method achieves AUROC improvements of up to 9.3\% over state-of-the-art techniques. Extensive ablation studies further validate the effectiveness of each component in our framework.
\end{abstract}

\section{Introduction}
\label{intro}
Large language models (LLMs) are rapidly being integrated into a variety of NLP tasks \citep{driess2023palm, bang2023multitask, zhong2023can, achiam2023gpt}. However, even widely adopted systems—such as ChatGPT \citep{openai2023chatgpt} and Gemini \citep{team2023gemini}—can generate content that is illogical or inconsistent with the given context, a phenomenon commonly referred to as “hallucination” \citep{Jisurvey}. Consequently, detecting hallucinations---that is, identifying inaccurate information generated by LLMs---has become a major focus in the literature.

Existing hallucination detection methods emphasize the semantic properties of text while minimizing reliance on lexical and syntactic features. The primary goal is to assess the accuracy of the generated information independent of its phrasing. When sampling multiple responses, semantically inconsistent answers to the same question indicate model uncertainty, which can signal hallucination. Building on the idea of leveraging semantic similarity and uncertainty across meaning distributions, \citet{kuhn2023semantic} introduced “Semantic Entropy,” an unsupervised method that identifies hallucinations by clustering generated responses based on semantic equivalence and subsequently computing the overall semantic entropy from the uncertainty within each cluster. While effective, this approach relies on Natural Language Inference (NLI) to determine semantic equivalence. NLI, however, struggles to capture the full range of semantic properties in text \citep{arakelyan-etal-2024-semantic} and is computationally intensive due to the large-scale transformer architectures underpinning NLI models \citep{percha2021}.

To address these limitations, we propose an optimized semantic clustering approach that leverages semantic similarity to compute semantic inconsistency over meanings using our novel measure, the Semantic INconsistency Index (SINdex). Our framework employs sentence embeddings to capture nuanced semantic properties in a high-dimensional space, followed by hierarchical clustering to efficiently group LLM responses. The resulting improvement in cluster homogeneity leads to more accurate uncertainty estimates and enhanced hallucination detection.

The primary contributions of this work are as follows:
\begin{itemize}
    \item We introduce a versatile black-box framework for automated hallucination detection across diverse LLMs, requiring no access to internal model states or external knowledge.
    \item We propose SINdex, a novel inconsistency measure that combines semantic cohesion and consistency to effectively detect hallucinations in LLM outputs.
    \item Scalability experiments demonstrate our framework’s superior efficiency, achieving a 60-fold speedup over state-of-the-art hallucination detection methods in large-scale settings (e.g., 200 generations).
    \item Our approach significantly enhances hallucination detection performance across a range of well-established open- and closed-book Question Answering (QA) datasets, including TriviaQA, NQ, SQuAD, and BioASQ.
\end{itemize}

The remainder of this paper is organized as follows. Section~\ref{sec:related_work} reviews related work on semantics in natural language generation. Section~\ref{method} presents our methodology and SINdex. Section~\ref{experiment} details our experimental setup, and Section~\ref{results} discusses our findings and ablation studies. Finally, Section~\ref{conclusion} concludes the paper and outlines future work.

\section{Background and Related Work}
\label{sec:related_work}
The deployment of LLMs in sensitive domains (e.g., medicine and law) is often constrained by their tendency to generate text that, while plausible, may lack factual grounding \citep{pal2023med, dahl2024large}. To address this challenge, researchers have developed a variety of methods that can be broadly classified into black-box, white-box, and gray-box approaches. 

Black-box approaches rely solely on the generated output, leveraging features such as text consistency and variability. For instance, \citet{manakul-etal-2023-selfcheckgpt} argue that an LLM’s knowledge about a concept can be inferred by sampling multiple responses; high consistency among responses suggests correctness, whereas significant divergence indicates potential hallucination. This output-centric perspective allows these methods to be readily applied to any LLM without internal modifications.  In contrast, white-box methods exploit the internal representations of LLMs (e.g., hidden layer activations) to detect and mitigate hallucinations \citep{burns2022discovering, li2024inference, azaria-mitchell-2023-internal}. Although these techniques often achieve improved performance over their black- and gray-box counterparts \citep{zhu2024pollmgraph}, the performance gains are marginal \citep{xiong2023can}. Moreover, accessing internal model states is impractical for users constrained by API limitations or when dealing with proprietary systems. Gray-box techniques bridge these approaches by harnessing token-level probabilities to derive additional metrics such as confidence scores or predictive uncertainty \citep{xiong2023can, xiao-wang-2021-hallucination, yuan2021bartscore}. While these methods offer a good trade-off between performance and accessibility, they typically overlook the underlying semantic structure of the generated text. Recent work by \citet{kuhn2023semantic} has demonstrated that incorporating text semantics---via the notion of \emph{semantic entropy}---can substantially enhance the detection of hallucinations.

\paragraph{Semantics in Natural Language Generation (NLG).}
Natural language is inherently multifaceted, with the same meaning often expressible through diverse syntactic forms. It is therefore critical to distinguish between syntax (the grammatical structure), lexical content (the choice of words), and semantics (the underlying meaning) \citep{Lyons_1995}. In the context of hallucination detection, a model’s semantic integrity is paramount; the reliability of generated responses hinges on accurately capturing the intended meaning, even when the surface form varies. 

\paragraph{Semantic Uncertainty and Entropy.}  
Building on this observation, \citet{kuhn2023semantic} proposed measuring model uncertainty by focusing on the semantic equivalence of sentences. For example, the sentences “rhinoviruses are the predominant cause of the common cold” and “the common cold is caused by rhinoviruses” differ syntactically yet convey the same information. Conventional token-based uncertainty measures may misinterpret such variations as significant, while a semantic entropy framework adjusts for these differences to reflect true uncertainty. Extending this idea, \citet{farquhar2024detecting} introduced a discrete semantic entropy measure that operates in a black-box setting, obviating the need for access to token probabilities.

\paragraph{Limitations of Current Semantic Approaches.}  
Despite the promise of semantic entropy, current methods often rely on bidirectional NLI models to perform semantic clustering. NLI models, which determine whether pairs of sentences entail or contradict each other, can struggle with the continuous and nuanced nature of semantic similarity \citep{naik-etal-2018-stress}. Moreover, such models frequently overemphasize lexical cues rather than deeper semantic relations \citep{arakelyan-etal-2024-semantic} and are computationally demanding due to their reliance on large-scale transformer architectures \citep{kabbara-cheung-2022-investigating}.

\section{Methodology}  
\label{method}

\begin{figure*}[ht]
\centering

\includegraphics[width=\textwidth]{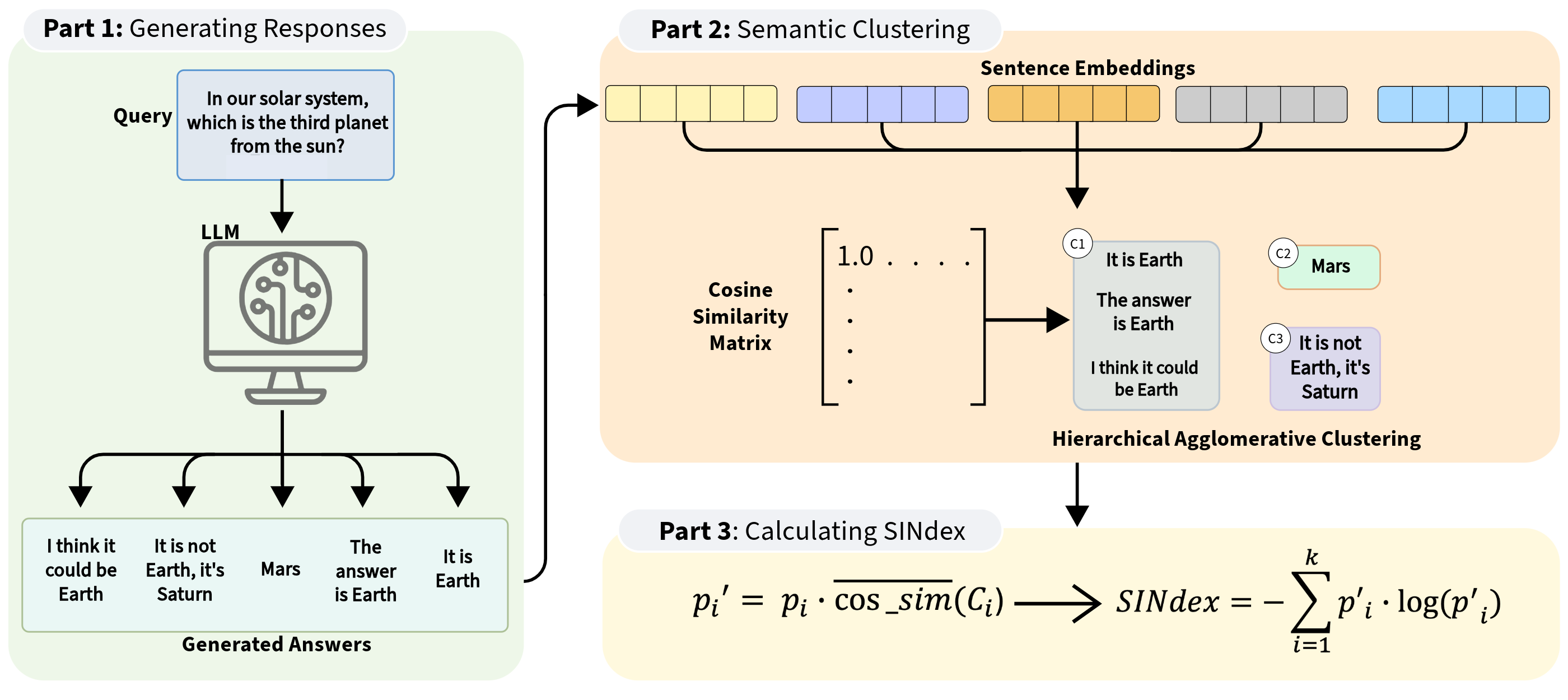}
\caption{Illustration of the proposed Natural Language Generation hallucination detection framework, which leverages an optimized semantic clustering approach to compute semantic entropy. \textbf{Part 1} involves generating multiple responses to the same question. \textbf{Part 2} processes these responses by computing sentence embeddings and clustering them via hierarchical agglomerative clustering. \textbf{Part 3} calculates the SINdex measure using an adjusted probability computation derived from the resulting clusters.}
\label{fig:method}
\end{figure*}

This section outlines our automated black-box method for hallucination detection in LLMs. The proposed framework consists of two components: (i) semantic clustering to group outputs with similar meanings, and (ii) SINdex, a novel measure that measures semantic inconsistency within these clusters as an indicator of model uncertainty. The semantic clustering component aggregates responses that share the same underlying meaning, regardless of syntactic or lexical differences, without relying on external prompts or manual intervention. The SINdex inconsistency measure quantifies coherence within these groups and flags potential hallucinations when inconsistencies arise. An overview of the methodology is provided in Fig. \ref{fig:method}.

\paragraph{Notation and Problem Statement.}\label{notation} Consider the problem of automatic hallucination detection in NLG, specifically for QA benchmarks. Let \(q\) denote a question used to prompt an LLM, and let \(g\) represent an LLM-generated response. To capture model uncertainty, the LLM is prompted \(P\) times, yielding a set of responses:
\begin{equation}
G = \{g_{1}, g_{2}, \ldots, g_{P}\}.
\end{equation}
For each \(g_i \in G\), a representative string \(s_i\) is constructed by concatenating the question \(q\), a separator token \([SEP]\), and the corresponding response \(g_i\):
\begin{equation}
s_i = q \circ [SEP] \circ g_i.
\end{equation}
A sentence embedding for each \(s_i\) is computed using a sentence similarity model \(Emb\), yielding
\begin{equation}
Emb(s_i) \in \mathbb{R}^{d},
\end{equation}
where \(d\) is the embedding dimension. These embeddings serve as the basis for semantic clustering and the subsequent computation of the SINdex measure, which quantifies semantic inconsistency to detect potential hallucination.

\subsection{Iterative Generation of Outputs}
The initial step consists of iteratively prompting the LLM \(P\) times with the query \(q\), thereby generating multiple independent responses. This approach guarantees that each output is produced without any influence from preceding responses.

\subsection{Semantic Clustering}
\paragraph{Generating Embeddings.} For each generated answer, the question \(q\) is concatenated with the response \(g_{i}\) using a separator token, yielding a string \(s_{i}\) that preserves the context of \(q\). A text embedding, \(Emb(s_{i})\), is then produced using a transformer-based model fine-tuned on a sentence similarity task. Cosine similarity is employed to measure the semantic similarity between embeddings, as defined in Equation~\ref{eq:cos_sim}:

\begingroup
\small
\begin{equation}
\mathrm{cos\_sim}\big( Emb(s_{i}), Emb(s_{j}) \big) = \frac{\langle Emb(s_{i}), Emb(s_{j}) \rangle}{\|Emb(s_{i})\| \cdot \|Emb(s_{j})\|}
\label{eq:cos_sim}
\end{equation}
\endgroup

Cosine similarity is a suitable measure for comparing semantic embeddings because it emphasizes the angle \(\theta\) between vectors rather than their magnitudes \citep{mikolov2013}. This focus on direction makes it particularly effective in high-dimensional spaces, where sparsity can obscure raw magnitude comparisons, as the emphasis on \(\theta\) captures semantic overlap regardless of dimensionality \citep{turney2010}. Additionally, the length-invariant normalization inherent in cosine similarity neutralizes variations in vector lengths, thereby isolating the true semantic relationship between embeddings.

\paragraph{Hierarchical Agglomerative Clustering.} Hierarchical agglomerative clustering is employed to partition the responses into an optimal number of groups. Initially, each embedding \(\{Emb(s_1), Emb(s_2), \ldots, Emb(s_P)\}\) forms its own cluster, denoted by \(C_1, C_2, \ldots, C_P\) where \(C_i = \{Emb(s_i)\}\). The algorithm then iteratively merges the closest clusters based on a distance function, \(dis(C_i, C_j)\), defined according to a chosen linkage criterion. In this work, a distance threshold of 0.05 is maintained throughout the experiments. 

Single linkage can inadvertently connect unrelated clusters, while complete linkage is overly sensitive to outliers \citep{RAMOSEMMENDORFER2021106990}. To address these limitations, average linkage is adopted as it provides a more balanced measure of similarity between clusters. The distance between embeddings \(s_i\) and \(s_j\) is defined as follows:

\begingroup
\small
\begin{equation}
\mathrm{dis}\big( Emb(s_i), Emb(s_j) \big)
= 1 - \mathrm{cos\_sim}\big( Emb(s_i), Emb(s_j) \big)
\label{eq:distance}
\end{equation}
\endgroup

The pseudocode for the clustering algorithm is provided in Appendix~\ref{algo}.

\begin{figure*}[ht]
    \centering
    \begin{subfigure}[b]{0.45\textwidth}
        \centering
        \includegraphics[width=\textwidth]{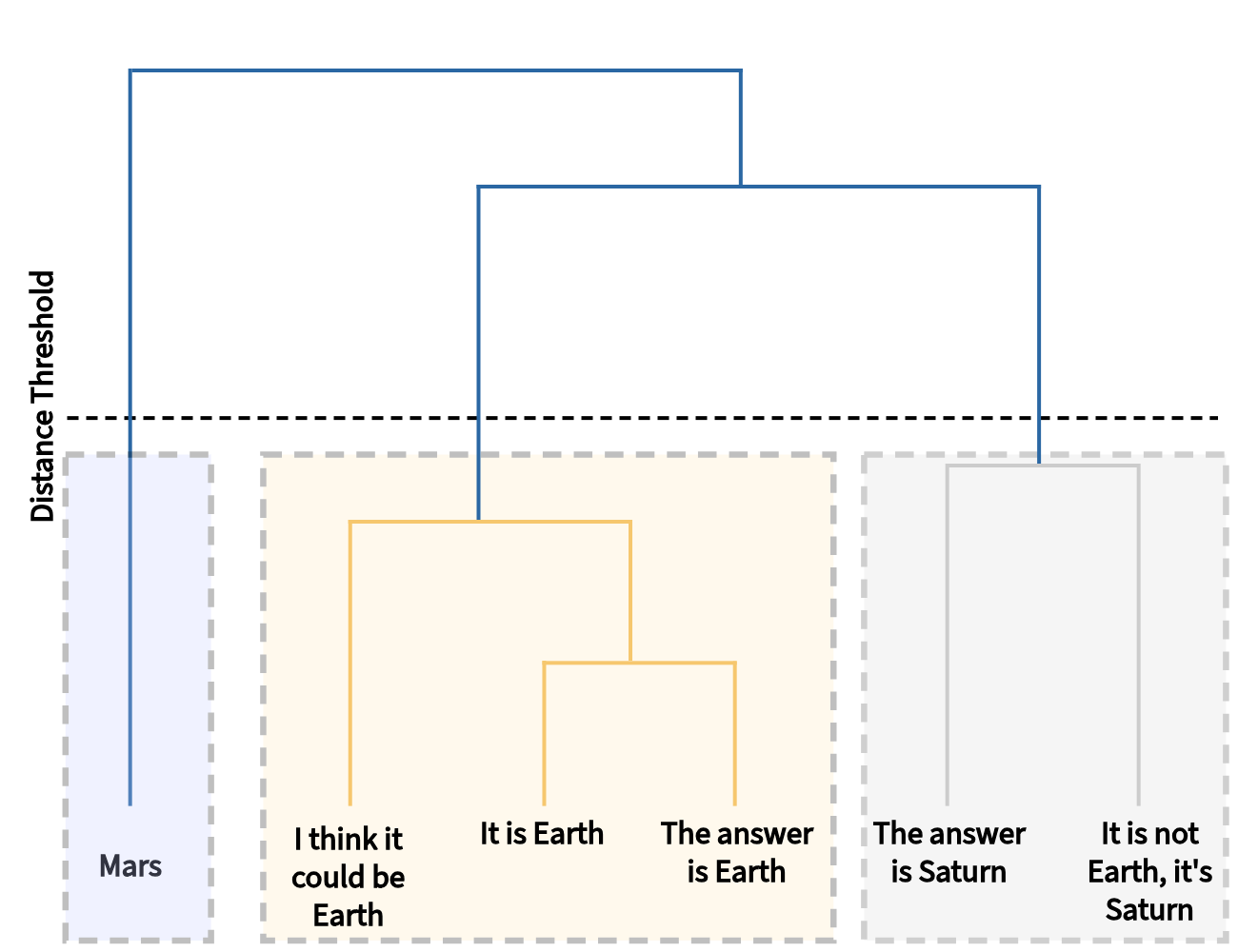}
        \caption{Our Semantic Clustering.}
        \label{fig:dendogram}
    \end{subfigure}
    \hfill
    \begin{subfigure}[b]{0.45\textwidth}
        \centering
        \includegraphics[width=\textwidth]{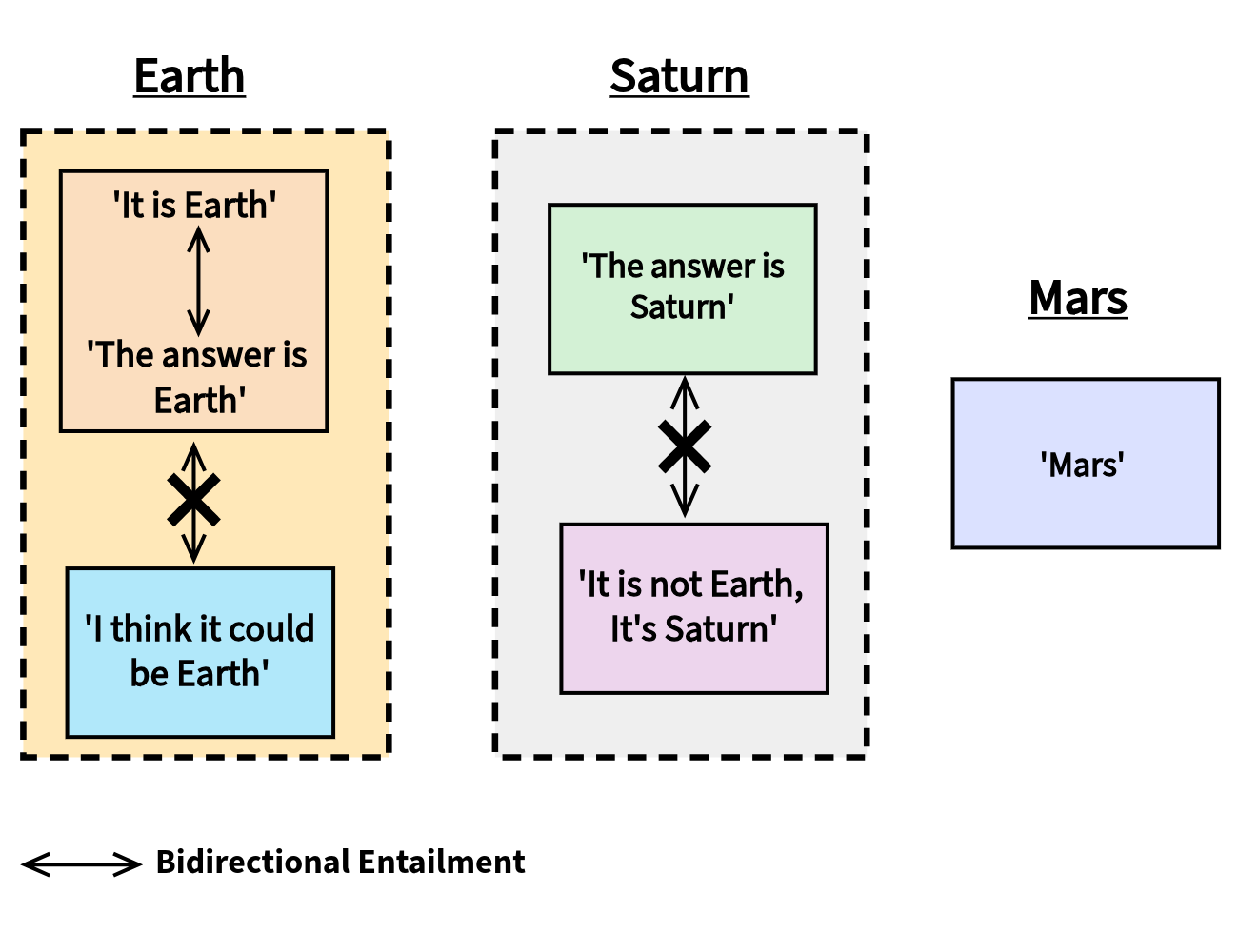}
        \caption{Bidirectional NLI Clustering.}
        \label{fig:nli_clustering}
    \end{subfigure}
    
\caption{Visualization of clusters obtained through hierarchical agglomerative clustering (left) and NLI-based clustering (right) for the same sample. Dashed squares denote the clusters identified by each method.}
    \label{fig:clustering_comparison}
\end{figure*}

\paragraph{Rationale Behind Hierarchical Agglomerative Clustering.}Hierarchical agglomerative clustering produces more homogeneous clusters than NLI-based bidirectional clustering. For example, consider the query 
$q$ = `In our solar system, which is the third planet from the sun?'
and the generated responses 
$G$ = [`It is Earth', `The answer is Saturn',`The answer is Earth', `Mars', `It is not Earth, it's Saturn', `I think it could be Earth']
Ideally, three clusters corresponding to \{\texttt{Earth}, \texttt{Saturn}, \texttt{Mars}\} should be formed. As shown in Fig.~\ref{fig:clustering_comparison}(a), agglomerative clustering yields three clusters, whereas bidirectional NLI clustering, depicted in Fig.~\ref{fig:clustering_comparison}(b), produces five clusters. Notably, ``I think it could be Earth'' is grouped with ``It is Earth'' and ``The answer is Earth'', and ``It is not Earth, it's Saturn'' is clustered with ``The answer is Saturn''. In contrast, bidirectional NLI clustering fails to merge these semantically similar responses because, while ``It is not Earth, it's Saturn'' entails ``The answer is Saturn'', the reverse is not true due to the absence of negation.

\paragraph{Complexity Analysis.}
Our  method comprises three main steps: (a) generating sentence embeddings, (b) computing pairwise cosine similarity between embeddings, and (c) clustering the generated embeddings. For the first step, assume each input question yields \(P\) answers. Each answer undergoes tokenization and a forward pass through a transformer model, with a computational cost of 
\[
O(P \cdot L^{2} \cdot d),
\]
where \(L\) denotes the number of tokens per answer and \(d\) is the embedding dimensionality. Next, computing pairwise cosine similarity requires \(\frac{P(P-1)}{2}\) comparisons; considering the embedding dimensionality, this step has a complexity of 
\[
O(P^{2} \cdot d).
\]
Finally, agglomerative clustering incurs a cost of 
\[
O(P^{2} \cdot \log P).
\]
Thus, the overall complexity is 
\[
O\Big(P \cdot L^{2} \cdot d\Big) + O\Big(P^{2}(d + \log P)\Big).
\]
\noindent\textbf{Scalability Analysis.}
To compare the scalability of the clustering approach with the NLI-based method, a runtime analysis was conducted across a varying number of generations. For this purpose, the NLI-based method was reimplemented using the DeBERTa-large model\footnote{\url{https://huggingface.co/microsoft/deberta-large-mnli}}, following the procedure detailed in \citet{kuhn2023semantic}. As shown in Appendix~\ref{app:runtime}, the results demonstrate that the clustering approach scales significantly better than the NLI-based method.

\subsection{SINdex: \textbf{S}emantic \textbf{IN}consistency Index for Hallucination Detection in LLMs}

The SINdex evaluates clustering quality by measuring coherence and distributional balance through an adjusted entropy formulation:

\begin{equation}
\text{SINdex} = -\sum_{i=1}^k p_i' \log(p_i'),
\end{equation}
where \( p_i' \) is the adjusted proportion of cluster \( i \), defined as:
\begin{equation}
p_i' = p_i \cdot \overline{\mathrm{cos\_sim}}(C_i).
\end{equation}
Here, \( p_i \) represents the proportion of points in cluster \( i \), satisfying:
\begin{equation}
\sum_{i=1}^k p_i = 1, \quad p_i \geq 0, \quad \forall i \in \{1, \dots, k\}.
\end{equation}
The term \( \overline{\mathrm{cos\_sim}}(C_i) \) denotes the average pairwise cosine similarity of points in cluster \( C_i \), explicitly defined as:

\begin{equation}
\overline{\mathrm{cos\_sim}}(C_i) = \frac{1}{\binom{|C_i|}{2}} \sum_{\substack{x, y \in C_i \\ x \neq y}} \mathrm{cos\_sim}(x, y),
\end{equation}
where \( |C_i| \) is the number of points in cluster \( i \),  \(\binom{|C_i|}{2}\) is the binomial coefficient, representing the number of unique pairs in the cluster, and
 \( \mathrm{cos\_sim}(x, y) \) computes the cosine similarity between points \( x \) and \( y \).\\
\noindent
\textbf{Theoretical Justification.}
Since points in a cluster lie within a cosine similarity of \([0, 1]\), the adjusted proportions \( p_i' \) satisfy:
\begin{equation}
0 \leq p_i' \leq p_i, \quad \forall i.
\end{equation}

Consequently, the adjusted proportion vector \( \mathbf{p}' = [p_1', p_2', \dots, p_k'] \) is \textbf{majorized} by the original proportion vector \( \mathbf{p} = [p_1, p_2, \dots, p_k] \). As Shannon entropy is Schur-concave \cite{entropy_concave}, it follows that:
\begin{equation}
H(\mathbf{p}') \geq H(\mathbf{p}),
\end{equation}
where \( H(\cdot) \) denotes entropy.

This relationship implies that the adjusted entropy \( H(\mathbf{p}') \) is always greater than or equal to the original entropy \( H(\mathbf{p}) \). Furthermore, \( H(\mathbf{p}') \) decreases monotonically with increasing intra-cluster coherence, as measured by \( \overline{\mathrm{cos\_sim}}(C_i) \). In the limiting case of perfect coherence (i.e., all pairwise cosine similarities are 1), the adjusted proportions reduce to the original proportions, yielding:
\begin{equation}
H(\mathbf{p}') = H(\mathbf{p}).
\end{equation}

\paragraph{Interpretation.}
The SINdex measure captures the semantic inconsistency of LLM responses to a query. A lower SINdex value indicates higher intra-cluster coherence and greater consistency in LLM responses by minimizing the dispersion of responses across semantic clusters while maximizing intra-cluster semantic coherence. Its entropy-based formulation ensures computational efficiency and straightforward applicability, making it particularly useful for evaluating clusters generated by LLMs.

\section{Experiments}
\label{experiment}

\paragraph{Data.}\label{data} 
Evaluation is conducted on four widely-used QA datasets: TriviaQA \citep{joshi-etal-2017-triviaqa}, a trivia-style closed-book dataset with concise answers; Natural Questions (NQ) \citep{kwiatkowski-etal-2019-natural}, containing Google search-derived questions with similarly brief responses; SQuAD \citep{rajpurkar-etal-2016-squad}, an open-book dataset featuring longer, descriptive answers; and BioASQ \citep{bioasq}, a life sciences dataset encompassing both binary (yes/no) and extended sentence answers. Representative samples from each dataset are provided in Appendix~\ref{app:data_samples}. As reported in the literature \citep{farquhar2024detecting}, we use a random sample of 400 questions from each dataset for our experiments. 

\paragraph{Models.} Experiments utilize state-of-the-art LMs, including Llama 2 \citep{Touvron2023Llama2O}, Mistral \citep{jiang2023mistral7b}, and Falcon \citep{almazrouei2023falconseriesopenlanguage}. The focus is on fine-tuned and instruction-tuned variants—namely, Llama-2-7b-chat, LLaMa-2-13b-chat, Falcon-7b-instruct, and Mistral-7b-instruct. To demonstrate broad applicability, no additional fine-tuning is performed; the pretrained models and their associated tokenizers available on Hugging Face are used directly.

\paragraph{Baselines.} Comparisons are made against four methods implemented by \citet{farquhar2024detecting}\footnote{\url{https://github.com/jlko/semantic_uncertainty}}. Specifically, the evaluation includes: \textbf{semantic entropy}; a supervised \textbf{embedding regression} approach \citep{kadavath2022languagemodelsmostlyknow} that predicts hallucinations using a regression model trained on LLM hidden states; \textbf{naive entropy}, which calculates entropy without accounting for semantic similarity among diverse lexicalizations of the same concept; and \textbf{p(true)} \citep{kadavath2022languagemodelsmostlyknow}, a few-shot prompt-based method for estimating the accuracy of LM outputs. Results from \textbf{SelfCheckGPT} \citep{manakul-etal-2023-selfcheckgpt} are also reported.


\paragraph{Automated Ground-Truth Label.} For each question, a canonical answer is generated by setting the model temperature to 0.1. Both the ground truth and the model responses are embedded using the \textit{all-MiniLM-L6-v2} model, selected for its effectiveness in capturing semantic similarity. A response is classified as accurate if the cosine similarity between its embedding and that of the ground truth exceeds 0.95; otherwise, it is flagged as a hallucination.



\section{Results} \label{results}
Table~\ref{tab:results} demonstrates that the approach consistently outperforms baseline methods across nearly all model-dataset combinations. In particular, relative to the state-of-the-art semantic entropy technique, improvements of up to 7.6\% on TriviaQA, 9.3\% on NQ, 9.1\% on SQuAD, and 4.8\% on BioASQ are observed.

The approach excels on datasets with short responses (TriviaQA, NQ, and SQuAD) by capturing subtle semantic differences in minimal inputs. Advanced sentence embeddings effectively reveal nuanced relationships among generated answers, yielding an entropy score that accurately indicates potential hallucinations.

Performance on BioASQ is notably higher for both this approach and the semantic entropy baseline, likely due to the binary nature of some responses (see Appendix~\ref{bioasq}), which simplifies the separation and clustering process compared to datasets with more varied responses.

\begin{table*}[t]
\centering
\footnotesize
\caption{AUROC for hallucination detection on open-form QA datasets across four representative LLMs. Bold values indicate the best performance per experiment.}
\renewcommand{\arraystretch}{1.3} 
\setlength{\tabcolsep}{5pt} 
\begin{tabular}{lcccccc}
\toprule
\multirow{2}{*}{\textbf{Model}} & \multirow{2}{*}{\textbf{Method}} & \multicolumn{4}{c}{\textbf{Dataset}} \\ 
\cmidrule(lr){3-6}
 & & TriviaQA & NQ & SQuAD & BioASQ \\ 
\midrule
\multirow{6}{*}{\textbf{Llama-2-7b-chat}} 
 & p(True)              & 0.642  & 0.646  & 0.607  & 0.786  \\
 & Embedding Regression & 0.631  & 0.578  & 0.621  & 0.714  \\
 & Naive Entropy        & 0.731  & 0.723  & 0.715  & 0.680  \\
 & Semantic Entropy     & 0.763  & 0.739  & 0.764  & 0.870  \\
 & SelfCheckGPT         & 0.807  & 0.764  & 0.592  & 0.708  \\
 & \textbf{SINdex}      & \textbf{0.821} & \textbf{0.833} & \textbf{0.827} & \textbf{0.940} \\ 
\midrule
\multirow{6}{*}{\textbf{Llama-2-13b-chat}} 
 & p(True)              & 0.788  & 0.731  & 0.711  & 0.773  \\
 & Embedding Regression & 0.695  & 0.698  & 0.592  & 0.732  \\
 & Naive Entropy        & 0.701  & 0.695  & 0.655  & 0.603  \\
 & Semantic Entropy     & 0.803  & 0.742  & 0.754  & 0.881  \\
 & SelfCheckGPT         & 0.782  & 0.714  & 0.648  & 0.656  \\
 & \textbf{SINdex}      & \textbf{0.830} & \textbf{0.761} & \textbf{0.846} & \textbf{0.939} \\ 
\midrule
\multirow{6}{*}{\textbf{falcon-7b-instruct}} 
 & p(True)              & 0.630  & 0.518  & 0.535  & 0.403  \\
 & Embedding Regression & 0.733  & 0.656  & 0.633  & 0.842  \\
 & Naive Entropy        & 0.767  & 0.732  & 0.649  & 0.697  \\
 & Semantic Entropy     & 0.786  & 0.736  & 0.710  & 0.861  \\
 & SelfCheckGPT         & 0.807  & 0.663  & 0.524  & 0.617  \\
 & \textbf{SINdex}      & \textbf{0.821} & \textbf{0.818} & \textbf{0.797} & \textbf{0.910} \\ 
\midrule
\multirow{6}{*}{\textbf{mistral-7b-instruct}} 
 & p(True)              & 0.758  & 0.730  & 0.643  & 0.757  \\
 & Embedding Regression & 0.681  & 0.598  & 0.615  & 0.797  \\
 & Naive Entropy        & 0.764  & 0.739  & 0.687  & 0.765  \\
 & Semantic Entropy     & 0.793  & 0.788  & 0.733  & 0.882  \\
 & SelfCheckGPT         & 0.814  & 0.675  & 0.486  & 0.601  \\
 & \textbf{SINdex}      & \textbf{0.870} & \textbf{0.808} & \textbf{0.770} & \textbf{0.928} \\ 
\bottomrule
\end{tabular}
\label{tab:results}
\end{table*}

\subsection{Ablation Studies} \label{sec:ablations}
A comprehensive ablation analysis is performed to identify the optimal hyperparameter settings, algorithmic components, and transformer configurations used in the experiments.

\begin{figure}[h]
    \centering
    \includegraphics[width=\columnwidth]{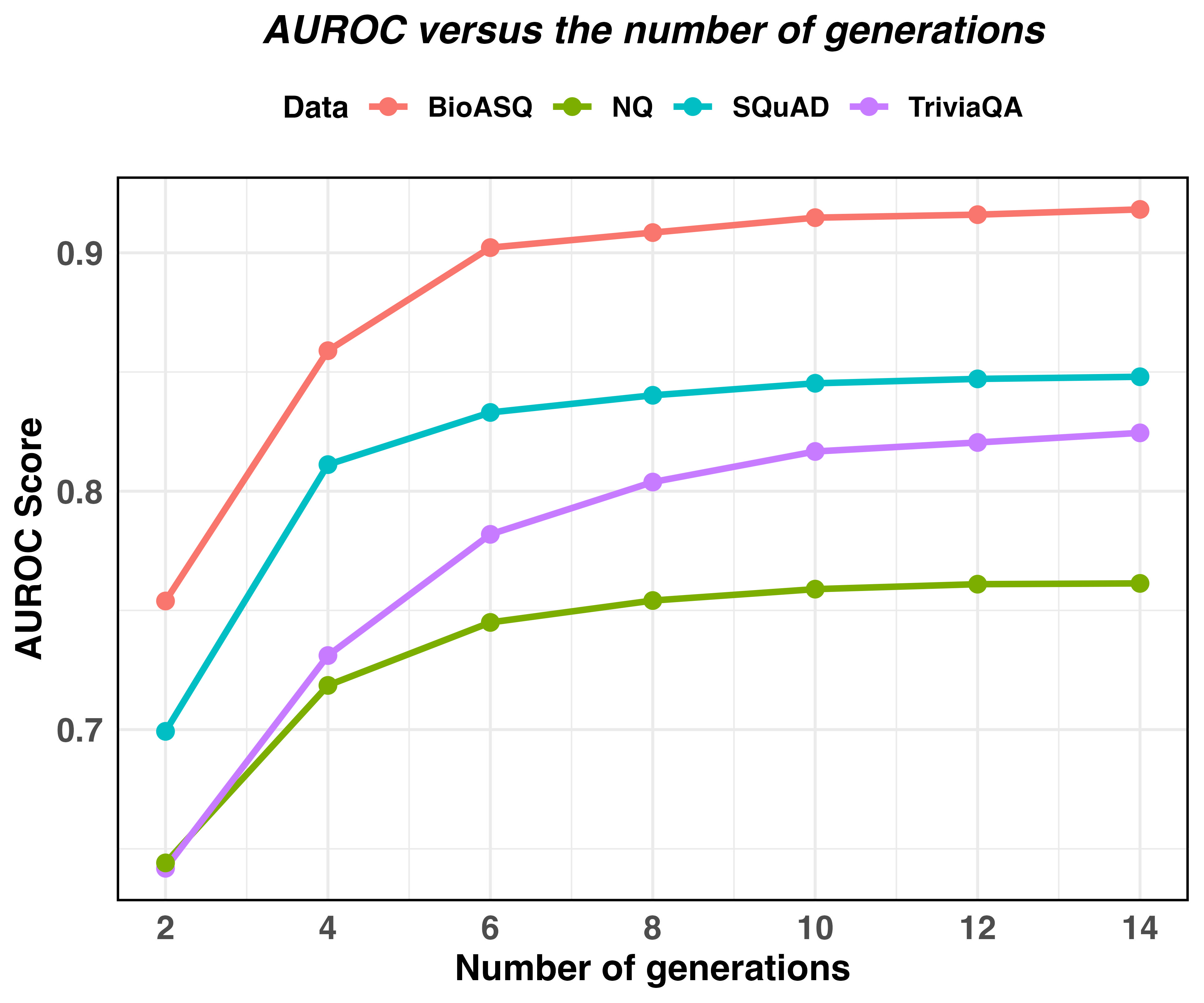}
    \caption{Ablation experiments evaluating the effect of the number of initial generations on LLaMa-2-13b-chat.}
    \label{fig:num_generations}
\end{figure}

\paragraph{Number of Generations.} The number of generations ($P$) is a critical hyperparameter affecting performance. Experiments were conducted with $P \in \{2, 4, 6, 8, 10, 12, 14\}$ across all four datasets to assess its impact on AUROC. 
Fig.~\ref{fig:num_generations} shows that AUROC values generally increase with an increase in $P$. However, when $P>10$, the increase is limited and the AUROC starts to level off. Consequently, we set $P=10$ through our experiments. Apart from achieving the best AUROC, a lower $P$ also reduces the inference costs associated with a higher number of generations.

\label{app1:cosine_sim_thresh}
\begin{figure}[h]
    \centering
    \includegraphics[width=\columnwidth]{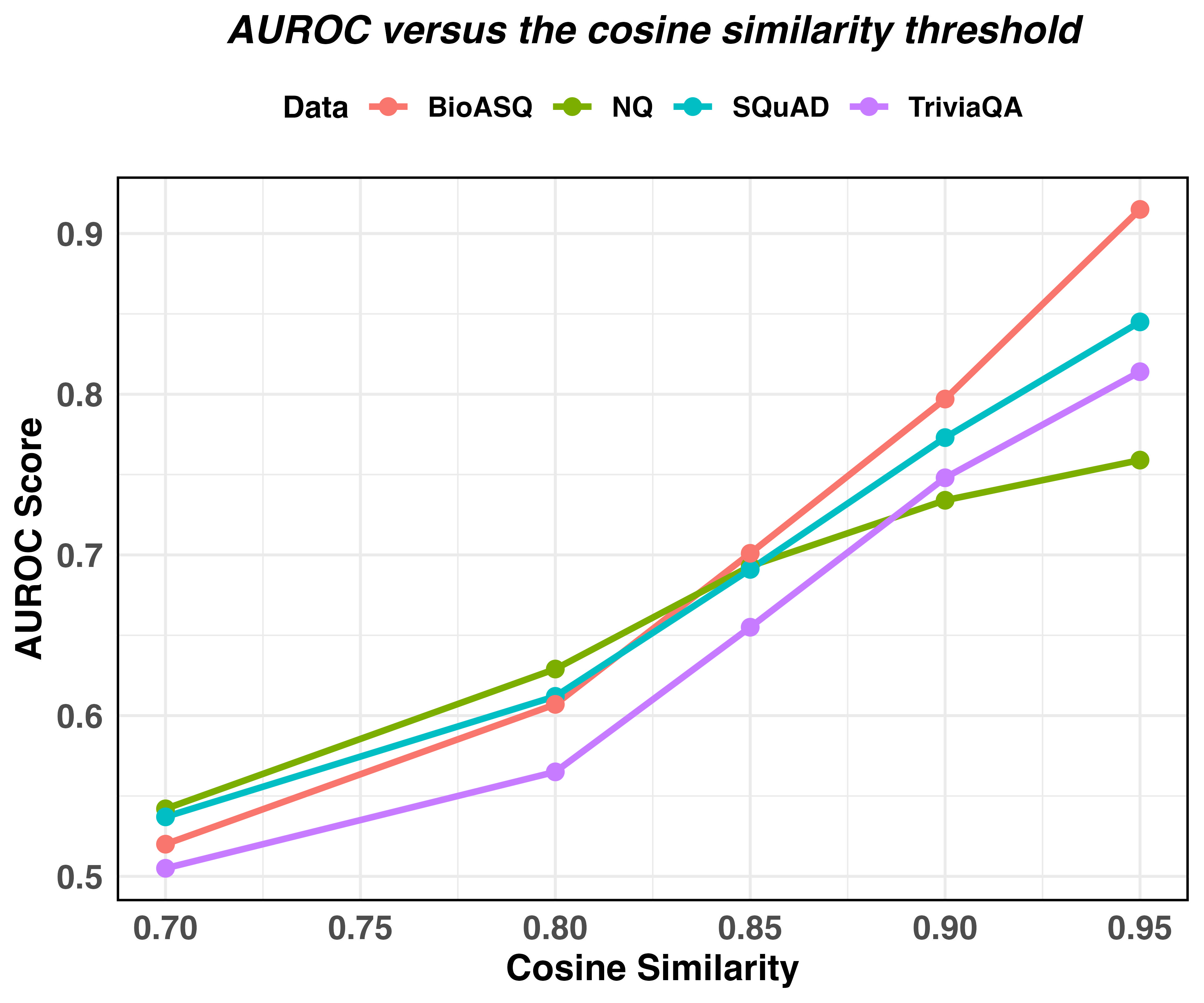}
    \caption{Ablation experiments on LLaMa-2-13b-chat evaluating the sensitivity of the cosine similarity threshold for semantic clustering across all datasets.}
    \label{fig:cosine_thresh}
\end{figure}

\paragraph{Cosine Similarity Threshold for Clustering.} 
Similarity thresholds in the range \(\{0.70, 0.80, 0.85, 0.90, 0.95\}\) were evaluated. As shown in Fig. ~\ref{fig:cosine_thresh}
, higher thresholds enhance clustering effectiveness, resulting in improved AUROC scores across all datasets. A threshold of 0.95 was selected for experiments; thresholds above 0.95 proved too rigid and degraded performance (see Appendix~\ref{app: cosine_sim_app} for results on TriviaQA).

\paragraph{Sentence Transformer Model for Semantic Similarity Clustering.} Several transformer models that produce semantically rich embeddings were evaluated for clustering. Experiments were conducted with popular models fine-tuned on sentence similarity tasks available on Hugging Face, including \textit{all-MiniLM-L6-v2}\footnote{\url{https://huggingface.co/sentence-transformers/all-MiniLM-L6-v2}}, \textit{all-mpnet-base-v2}\footnote{\url{https://huggingface.co/sentence-transformers/all-mpnet-base-v2}}, \textit{Alibaba-NLP/gte-large-en-v1.5}\footnote{\url{https://huggingface.co/Alibaba-NLP/gte-large-en-v1.5}}, and \textit{paraphrase-multilingual-MiniLM-L12-v2}\footnote{\url{https://huggingface.co/sentence-transformers/paraphrase-multilingual-MiniLM-L12-v2}}. Results on LLaMa-2-13b-chat and the TriviaQA dataset (see Appendix~\ref{app:sentence_sim}) show that \textit{all-MiniLM-L6-v2} achieves the highest AUROC scores (Fig.~\ref{fig:sem_modelA}) while also offering superior runtime efficiency (Fig.~\ref{fig:sem_modelB}).

\paragraph{Clustering Algorithm.} To identify the optimal clustering algorithm based on cosine similarity between embeddings, experiments were conducted comparing DBSCAN and OPTICS with Agglomerative Hierarchical Clustering. As shown in Appendix~\ref{app:clustering_algorithm}, Agglomerative Hierarchical Clustering outperforms the other methods on LLaMa-2-13b-chat and TriviaQA.

\section{Conclusion}\label{conclusion}
This work presents a scalable, uncertainty-based semantic clustering framework for detecting hallucinations in LLM outputs, incorporating the novel SINdex inconsistency measure. Leveraging sentence embeddings and hierarchical agglomerative clustering, the framework produces homogeneous clusters and refined inconsistency scores that robustly identify hallucinations across diverse QA datasets. Experimental results show that SINdex surpasses state-of-the-art methods, thereby enhancing the reliability and trustworthiness of LLM outputs. Future work will extend this framework to additional NLG tasks and further optimize its efficiency for large-scale, real-world applications.

\section{Limitations}
Despite promising experimental results, several limitations remain. First, the framework relies on the semantic properties of LLM outputs to assess hallucination likelihood. While effective for textual data, this reliance may not generalize to non-textual domains such as mathematical expressions or code, where semantic representations differ markedly. Addressing hallucinations in these contexts is an important direction for future work.

Second, ablation studies (Section~\ref{sec:ablations}) indicate that a fixed cosine similarity threshold of 0.95 yields optimal clustering performance on the evaluated datasets. However, a static threshold may lead to over- or under-clustering in other settings, potentially compromising cluster accuracy. Future research will explore dynamic, data-driven thresholding techniques to improve the flexibility and generalizability of the semantic clustering approach.


\appendix

\section{Runtime Analysis}\label{app:runtime}

\begin{figure}[htbp]
    \centering    
    \includegraphics[width=0.9\columnwidth]{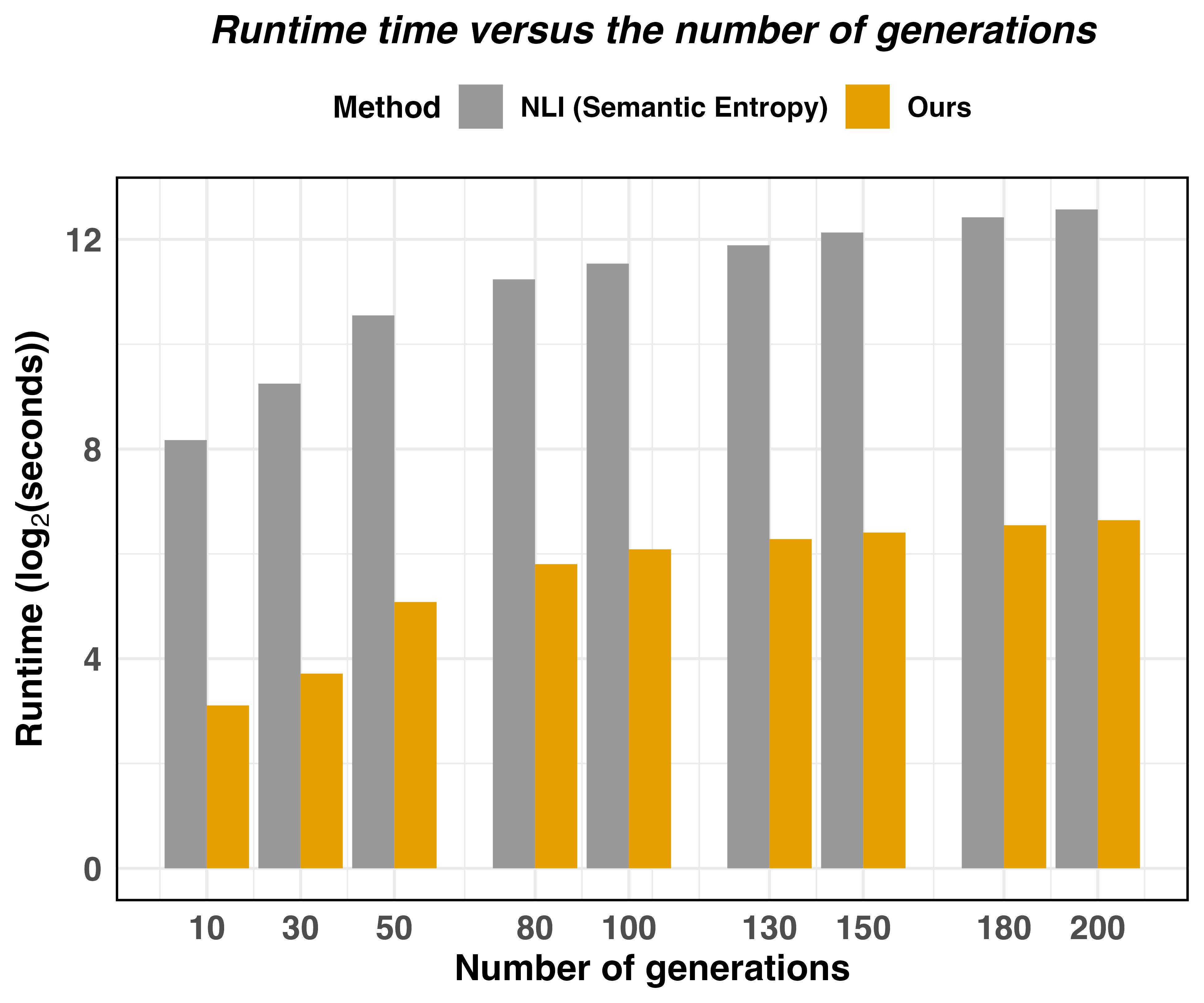}
    \caption{Runtime Analysis of NLI and agglomerative clustering over varying number of generations.}
    \label{fig:scalability}
 \end{figure}

Runtime scalability analysis comparing NLI vs. agglomerative clustering is presented in Fig. \ref{fig:scalability}. 
 
\section{Samples from QA Datasets}
\label{app:data_samples}

\subsection{TriviaQA}
\textbf{Question:} What was the name of the Oscar-winning song performed by Audrey Hepburn in `Breakfast at Tiffany's'?\\
\textbf{Answer:} Moon River\\
\rule{0.5\textwidth}{0.2mm}\\
\textbf{Question:} Late English criminal Bruce Reynolds masterminded which infamous robbery, which he later referred to as his ‘Sistine Chapel ceiling’? \\
\textbf{Answer:} Great Train Robbery

\subsection{NQ}
\textbf{Question:} Who is the actress that plays Aurora in Maleficent?\\
\textbf{Answer:} Elle Fanning
\\\rule{0.5\textwidth}{0.2mm}\\
\textbf{Question:} Who did Rome fight against in the Punic Wars?\\
\textbf{Answer:} Carthage

\subsection{SQuAD}
\textbf{Context:} The university is the major seat of the Congregation of Holy Cross (albeit not its official headquarters, which are in Rome). Its main seminary, Moreau Seminary, is located on the campus across St. Joseph lake from the Main Building. Old College, the oldest building on campus and located near the shore of St. Mary lake, houses undergraduate seminarians. Retired priests and brothers reside in Fatima House (a former retreat center), Holy Cross House, as well as Columba Hall near the Grotto. The university through the Moreau Seminary has ties to theologian Frederick Buechner. While not Catholic, Buechner has praised writers from Notre Dame and Moreau Seminary created a Buechner Prize for Preaching.
\\\rule{0.3\textwidth}{0.2mm}\\
\textbf{Question:} Which prize did Frederick Buechner create?\\
\textbf{Answer:} Buechner Prize for Preaching\\
\rule{0.5\textwidth}{0.2mm}\\
\textbf{Context:} All of Notre Dame's undergraduate students are a part of one of the five undergraduate colleges at the school or are in the First Year of Studies program. The First Year of Studies program was established in 1962 to guide incoming freshmen in their first year at the school before they have declared a major. Each student is given an academic advisor from the program who helps them to choose classes that give them exposure to any major in which they are interested. The program also includes a Learning Resource Center which provides time management, collaborative learning, and subject tutoring. This program has been recognized previously, by U.S. News \& World Report, as outstanding.
\\\rule{0.3\textwidth}{0.2mm}\\
\textbf{Question:} What was created at Notre Dame in 1962 to assist first year students?\\
\textbf{Answer:} The First Year of Studies program

\subsection{BioASQ}
\label{bioasq}
\textbf{Question:} What is the Daughterless gene?\\
\textbf{Answer:} The daughterless (da) gene in Drosophila encodes a broadly expressed transcriptional regulator whose specific functions in the control of sex determination and neurogenesis have been extensively examined.
\\
\rule{0.5\textwidth}{0.2mm}\\
\textbf{Question:} Is the FIP virus thought to be a mutated strain for the Feline enteric Coronavirus?\\
\textbf{Answer:} Yes

\section{Clustering Algorithm Psuedocode}\label{algo}

We show the semantic clustering pseudocode in Algorithm \ref{alg:embedding_clustering_avg}.
\begin{algorithm}
\caption{Clustering Algorithm with Average Distance}
\label{alg:embedding_clustering_avg}
\KwIn{set of sequences $S = \{s_{1}, s_{2}, \dots, s_{P}\}$; embedding model $Emb$; distance threshold $thresh$}
\KwOut{Set of clusters $C$}

Initialize empty set of clusters $C = \{\}$\;
\ForEach{sequence $s_{i} \in S$}{ 
    Compute embedding $ Emb(s_{i})$\; 
}
\ForEach{sequence $s_{i} \in S$}{
    Initialize a new cluster $c_{i} = \{s_{i}\}$\;
    \ForEach{cluster $c \in C$}{
        Initialize cumulative distance $total\_dis = 0$\;
        \ForEach{sequence $s^{(c)} \in c$}{
            Retrieve embedding $\mathbf{Emb}^{(c)} = Emb(s^{(c)})$\;
            Compute cosine similarity:
            \[
            \text{cos\_sim} = \frac{\langle \mathbf{Emb}(s_{i}), \mathbf{Emb}^{(c)} \rangle}{\|\mathbf{Emb}(s_{i})\| \cdot \|\mathbf{Emb}^{(c)}\|}
            \]
            Compute distance: $dis = 1 - \text{cos\_sim}$\;
            Accumulate the distance: $total\_dis \leftarrow total\_dis + dis$\;
        }
        Compute average distance (average linkage):
        \[
        \text{avg\_dis} = \frac{total\_dis}{|c|}
        \]
        \If{$\text{avg\_dis} \leq \text{thresh}$}{
            Merge $s_{i}$ into cluster $c$: 
            $c \leftarrow c \cup \{s_{i}\}$\;
            \textbf{break} (from the inner loop)\;
        }
    }
}
\Return{clusters $C$}\;
\end{algorithm}

\section{Implementation details} \label{imp}
We use Hugging Face to access transformer models and most datasets throughout the experiments. For BioASQ, we use the training dataset from Task B in the 2023 BioASQ challenge\footnote{http://participants-area.bioasq.org/datasets/}. 
Primary hyper-parameters to consider are: number of generations ($P$), which we set to $P=10$, generated by setting the model temperature to $1.0$, to keep it consistent with the baselines. Additionally, for automatic semantic clustering, we use the \textit{all-MiniLM-L6-v2} model to generate embeddings, and a cosine similarity threshold of $0.95$ (distance of $0.05$) for clustering.

\section{Ablation Results}
\label{app:ablation_appendix}

\subsection{Higher Cosine Similarity Threshold Reduces AUROC} 
\label{app: cosine_sim_app}
\begin{figure}[h]
    \centering
        \centering
        \includegraphics[width=0.9\columnwidth]{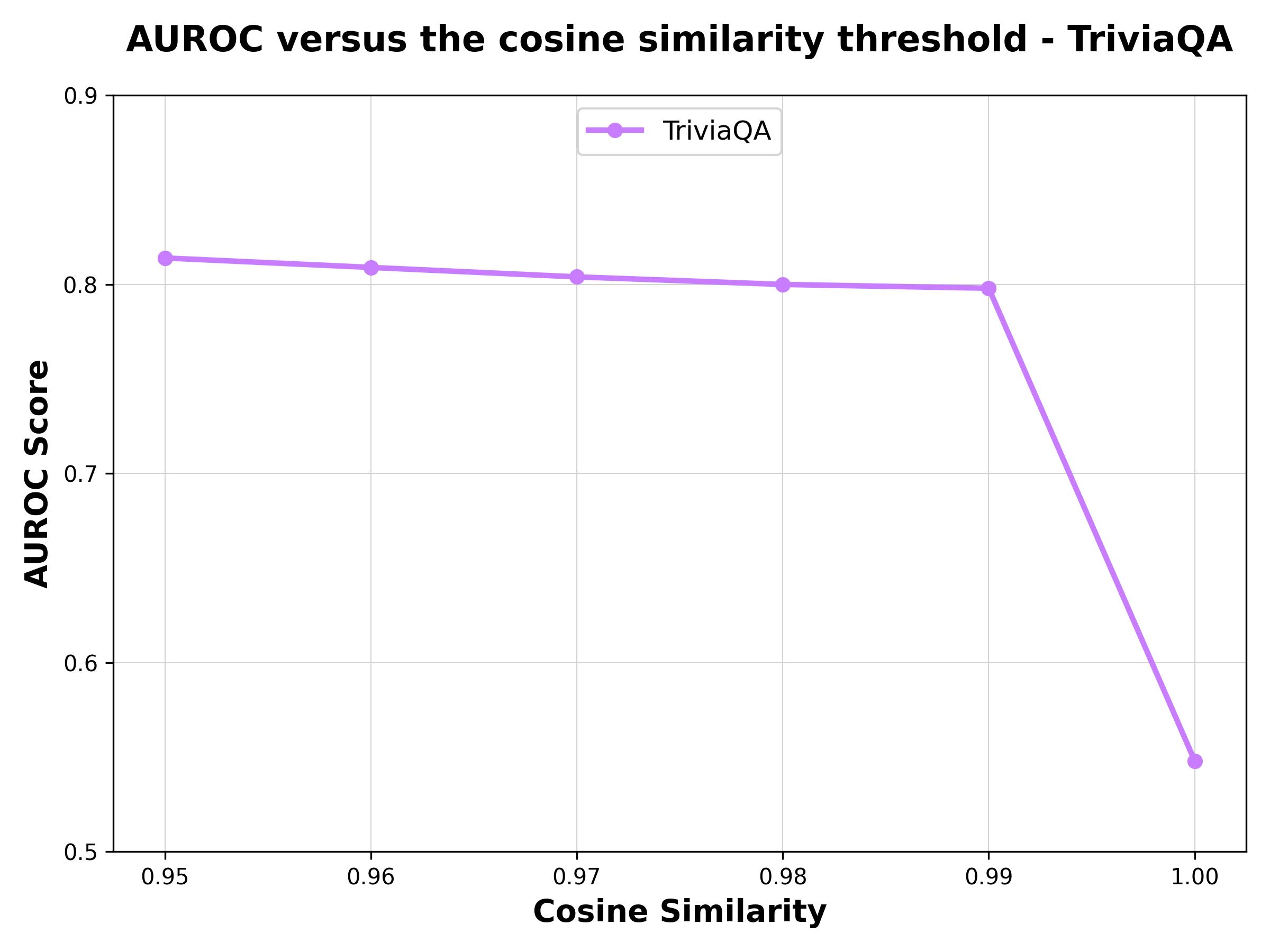}
        \caption{Variation in AUROC as a function of cosine similarity cutoff. The plot is generated with LLaMa-2-13b-chat on TriviaQA. The plot demonstrate the sensitivity of cosine similarity threshold used for semantic clustering.}
    \label{fig:cosine_thresh_appendix}
\end{figure}

Fig.~\ref{fig:cosine_thresh_appendix} shows the AUROC score on the TriviaQA dataset. Using a stringent similarity cutoff ($>0.95$) forces only highly similar embeddings to be clustered together-this reduces the scope for clustering semantically similar sentences which could be differently phrased.

\subsection{Sentence Similarity Models Ablation} \label{app:sentence_sim}
\begin{figure}[h]
    \centering
    \begin{subfigure}{\columnwidth}
        \centering
        \includegraphics[width=0.9\columnwidth]{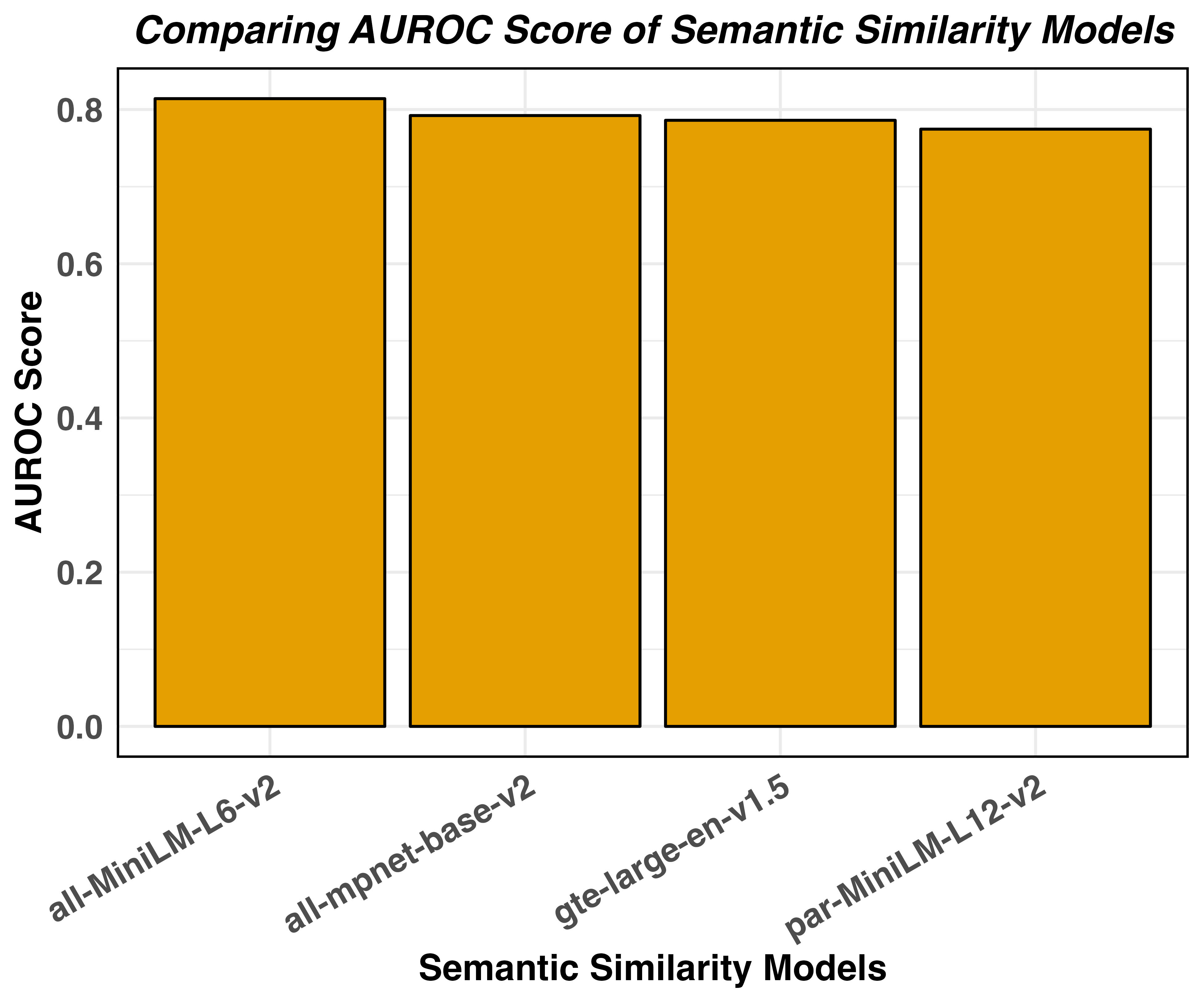}
        \caption{AUROC results for different models}
        \label{fig:sem_modelA}
    \end{subfigure}
    
    \vspace{0.5cm}  

    \begin{subfigure}{\columnwidth}
        \centering
        \includegraphics[width=\columnwidth]{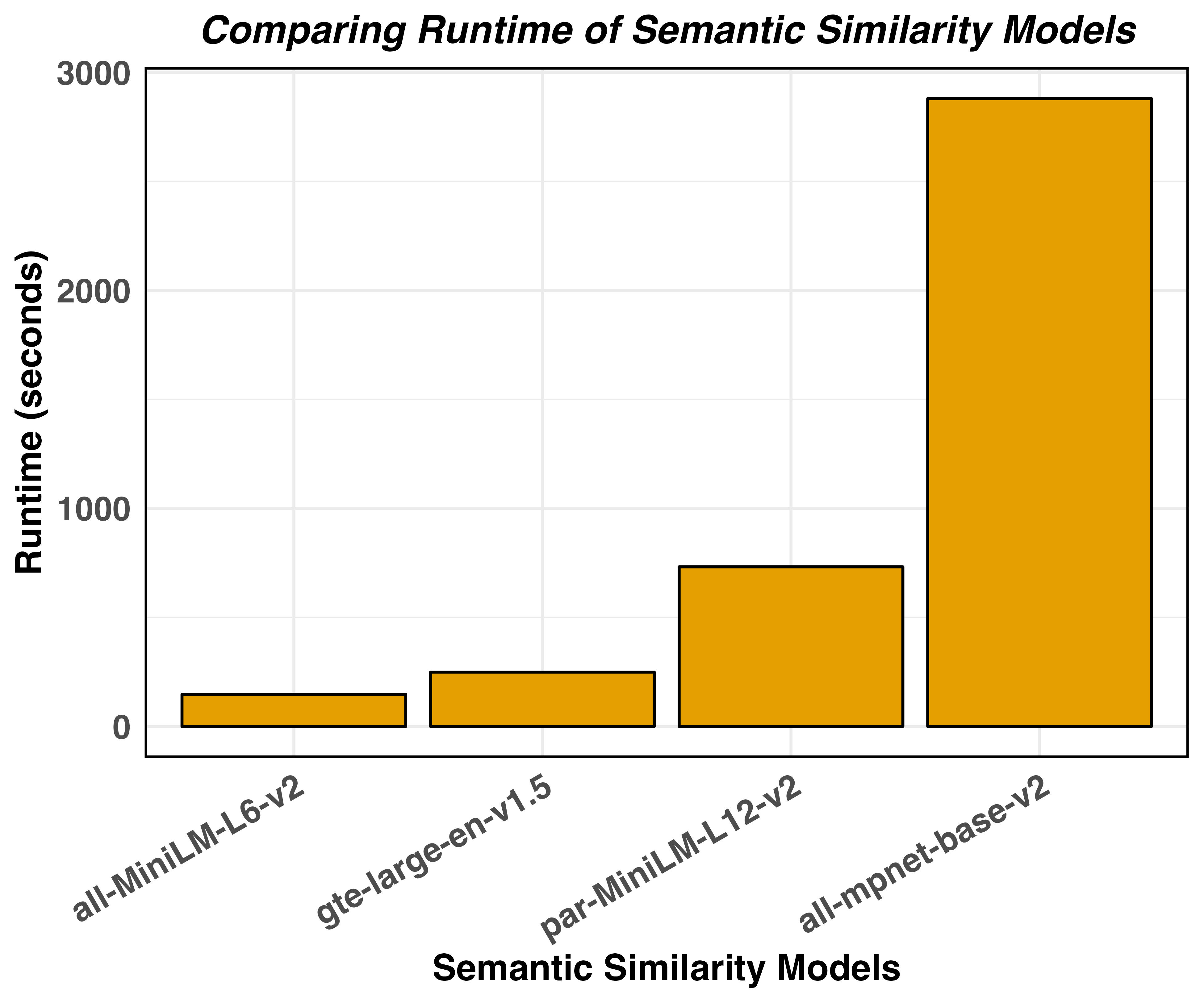}
        \caption{Runtime analysis for embedding generation}
        \label{fig:sem_modelB}
    \end{subfigure}

    \caption{(a) AUROC results when using different sentence similarity models. (b) Runtime analysis for generating embeddings using each model.}
    \label{fig:sem_model}
\end{figure}

Fig. \ref{fig:sem_model} illustrates ablation experiment results for different sentence similarity models. 

\subsection{Clustering Algorithms Ablation}
\label{app:clustering_algorithm}
\begin{figure}[!htbp]
    \centering
        \centering
        \includegraphics[width=0.9\columnwidth]{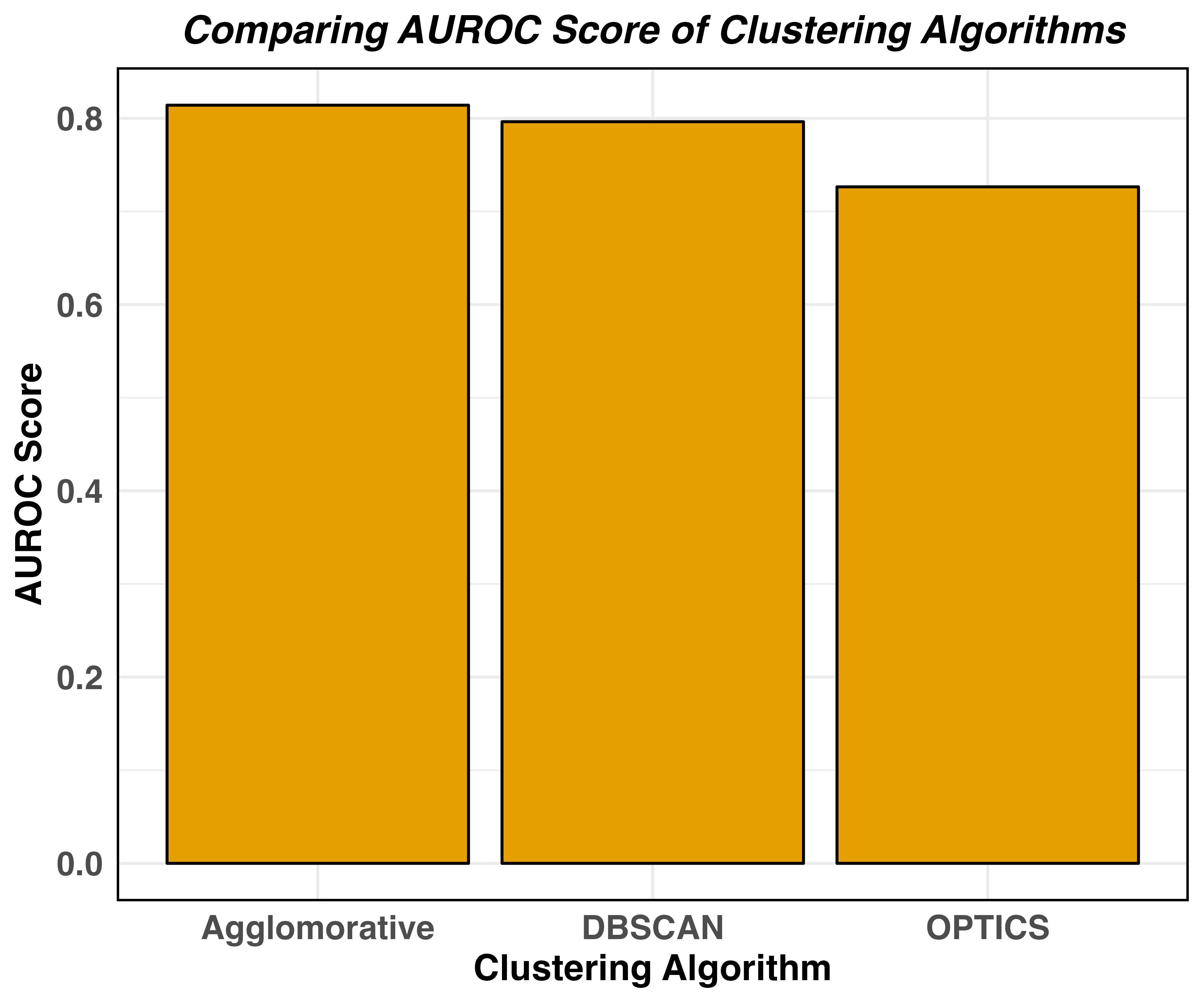}
        \caption{Comparison of AUROC obtained with clustering algorithms.}
    \label{fig:clustering_algorithm}
\end{figure}

As shown in Fig. \ref{fig:clustering_algorithm}, when experimenting with the LLaMa-2-13b-chat and TriviaQA dataset, we achieved AUROC scores of 0.796, 0.726, and 0.814, respectively. In this case, clustering achieves optimal performance, while detection performance shows a slight decline with the use of other clustering algorithms. 

\end{document}